\documentclass[letterpaper]{article} % DO NOT CHANGE THIS
\usepackage{aaai2027} % arXiv (non-anonymous) version
\usepackage[hyphens]{url} % DO NOT CHANGE THIS
\usepackage{graphicx} % DO NOT CHANGE THIS
\usepackage{natbib} % DO NOT CHANGE THIS AND DO NOT ADD ANY OPTIONS TO IT
\usepackage{caption} % DO NOT CHANGE THIS AND DO NOT ADD ANY OPTIONS TO IT
\usepackage{booktabs}
\usepackage{amsmath}

\title{WorldDynCache: Risk-Controlled Latent Dynamics Approximation for\\ Diffusion World Models}

\author{
    Leyang Chen,
    Junyi Wu,
    Shaoqiu Zhang,
    Yulun Zhang\thanks{Corresponding author.}
}
\affiliations{
    Shanghai Jiao Tong University
}

\begin{document}

\maketitle

\begin{abstract}
Diffusion world models generate high-quality futures, but repeated transformer evaluations make inference prohibitively slow. Existing caches reuse intermediate features, selectively update tokens, or reuse and extrapolate denoising outputs according to local drift or short native-space histories. These criteria can miss both approximation-induced latent transition defects that accumulate across skipped steps and phase- or condition-dependent changes in the direction of latent evolution. We propose WorldDynCache, a risk-controlled latent dynamics approximation framework with two core components. First, a lightweight latent-transition risk estimator tracks the accumulated future impact of approximation defects and calibrates its predictions against counterfactual defects observed at exact anchors. Second, a condition- and phase-aware lifted latent surrogate approximates latent evolution without extra transformer evaluations. On HunyuanVoyager-13B and Aether-5B, WorldDynCache achieves $4.92\times$ and $2.15\times$ speedups, respectively, while attaining the best generation quality among the compared caching methods across WorldScore, PSNR, SSIM, and LPIPS.
\end{abstract}

\section{Introduction}

Diffusion video world models~\citep{huang2025voyager,zhu2025aether} extend generative video models from unconstrained synthesis to conditional prediction of an environment's visual evolution. Given observed context, camera trajectories, and geometric cues, they generate spatially and temporally consistent views for controllable scene exploration and visual simulation~\citep{zhu2025aether,huang2025voyager}. This capability is especially valuable when explicit scene models or exhaustive capture of an environment are unavailable~\citep{zhu2025aether,huang2025voyager}.

This capability, however, comes with substantial inference cost~\citep{akbari2026flash}. A diffusion world model evolves a latent world state through many denoising steps, each of which typically invokes a large transformer~\citep{peebles2023dit,zhu2025aether,huang2025voyager}:
\[
    z_{t-1}
    =
    \Psi_t\!\left(z_t,F_\theta(z_t,t,c)\right),
\]
where $F_\theta$ denotes the learned denoising model, $\Psi_t$ is the scheduler transition, and $c$ denotes the model conditions. High-dimensional spatio-temporal tokens, large model capacity, and repeated full-model evaluations make the transformer the dominant computational cost of a rollout~\citep{liu2025teacache,feng2026worldcache}. This latency limits applications that require rapid generation or repeated prediction.

\begin{figure}[t!]
\centering
\includegraphics[width=\columnwidth]{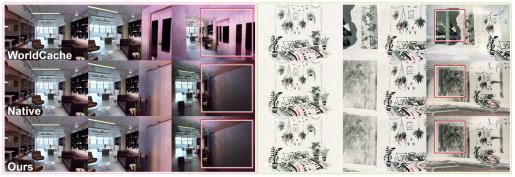}
\caption{Visualization comparison on two  representative WorldScore cases. Each row compares WorldCache, native output (Native), and our method (Ours). The red boxes highlight representative regions in which WorldDynCache better preserves native output than worldcache.}
\label{fig:worlddyncache-motivation}
\end{figure}

Caching methods reduce this cost by exploiting redundancy across neighboring denoising steps. Feature- and block-level methods reuse intermediate activations~\citep{ma2024deepcache,wimbauer2024blockcaching}; token-level methods selectively update predictable tokens~\citep{zou2024toca,zou2024duca}; output-level methods reuse or extrapolate denoising outputs and transformation vectors~\citep{liu2025teacache,zhou2025easycache}; and trajectory methods forecast future features from recent histories~\citep{liu2025taylorseer,feng2026hicache}. For diffusion world models, WorldCache further selects token predictors according to local token dynamics and resumes exact computation when accumulated token drift becomes large~\citep{feng2026worldcache}. Despite different prediction spaces, these methods share a local cacheability view: approximation is safe when the current representation remains predictable from nearby observations. 

\begin{figure}[t!]
\centering
\includegraphics[width=\columnwidth]{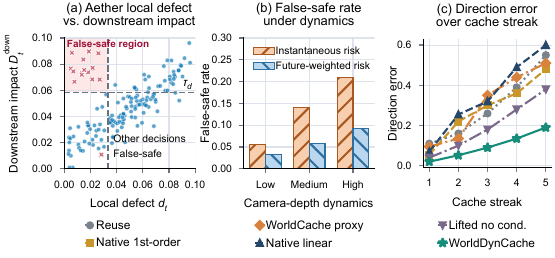}
\caption{Latent-transition safety diagnostics for WorldDynCache.
(a) Local transition defect $d_t$ versus downstream impact
$D_t^{\mathrm{down}}$, using thresholds
$\tau_D/\tau_d=0.060/0.025$ for Aether and $0.080/0.030$ for
Voyager.
(b) Instantaneous/future-weighted false-safe rates of
$5.2\%/3.8\%$, $14.6\%/5.4\%$, and $20.9\%/9.5\%$ in the
Low/Medium/High camera--depth dynamics bins.
(c) Direction errors of different approximation methods over
increasing cache-streak lengths. Full experimental details are
provided in the supplementary material.}
\label{fig:transition-safety-gap}
\end{figure}

This local view is incomplete for diffusion world model. Once an approximate result  is committed, its error enters the next latent state and is subsequently evolved by later denoising transitions. Thus, a small defect at the current step does not necessarily imply limited degradation over the remaining rollout. Moreover, the displacement observed in a short native-space history may cease to be informative when the denoising phase, camera motion, or scene geometry changes the direction of latent evolution. Based on carefully designed experiments, Figure~\ref{fig:transition-safety-gap} evaluates these two gaps: the mismatch between instantaneous transition defect and downstream degradation, and the loss of transition-direction accuracy over consecutive surrogate steps. These observations suggest that skipped computation should be treated as an approximation of the composed diffusion latent transition, whose reliability depends on both accumulated approximation risk and the current phase and conditions.

Based on this insight, we propose \emph{WorldDynCache}, a risk-controlled latent-transition approximation framework. Instead of first approximating a denoising output and then applying the scheduler, WorldDynCache directly constructs a candidate next latent, using exact transitions previously observed under compatible latent, scheduler-phase, camera, and geometry conditions. The candidate is constructed in a condition-augmented lifted representation using locally retrieved transition increments, following the observable-based modeling principle of Koopman methods without learning or identifying a global Koopman operator. A nonlinear memory-based observation then reconstructs the candidate in the original latent space.

WorldDynCache further determines whether this candidate should be committed. Exact steps expose counterfactual next-latent defects, which are used to calibrate an inexpensive online defect proxy. The controller increases the estimated risk when the current denoising phase or condition changes indicate greater sensitivity, accumulates this empirical risk across consecutive surrogate transitions, and falls back to the original transformer transition when the candidate is unsupported or its accumulated risk is excessive. Thus, the surrogate determines how to approximate the next latent, while the controller determines when that approximation is sufficiently reliable. Both components use cached exact anchors and lightweight condition summaries, require no auxiliary training, and introduce no additional transformer evaluation when a surrogate transition is accepted.
As shown in Figure~\ref{fig:worlddyncache-motivation}, WorldDynCache significantly outperforms previous sota method in representing the native output.

This paper makes three contributions:
\begin{itemize}
    \item We identify and empirically characterize two limitations of local cacheability criteria in diffusion world models: instantaneous transition defects can underestimate downstream rollout degradation, and native-space extrapolation can lose transition-direction accuracy as conditions and denoising phases evolve.

    \item We introduce WorldDynCache, which combines a condition- and phase-aware latent-transition surrogate with an online-calibrated risk controller. The surrogate directly approximates the composed next-latent transition, while the controller accumulates a condition-sensitive empirical risk and selectively falls back to exact transformer evaluation, without auxiliary training or additional transformer computation on accepted surrogate steps.

    \item We evaluate WorldDynCache on HunyuanVoyager-13B and Aether-5B world generation and on Aether 3D reconstruction. Under the evaluated protocols, it achieves $4.92\times$ and $2.15\times$ generation speedups while providing the strongest WorldScore and native-fidelity results among the evaluated accelerated methods, and reaches $3.42\times$ speedup on reconstruction with depth and pose metrics that outperform the compared caching baselines.
\end{itemize}

\section{Related Work}

\subsection{Diffusion Models and Video World Models}

Diffusion models generate samples through iterative latent transitions~\citep{ho2020ddpm,rombach2022ldm}. Efficient samplers reduce denoiser evaluations using non-Markovian trajectories, diffusion-specific ODE solvers, redesigned noise parameterizations, or progressive distillation~\citep{song2020ddim,lu2022dpmsolver,karras2022edm,salimans2022progressive}. Unlike caching, these approaches shorten or learn the sampling trajectory. Sequential coupling can propagate local prediction errors into later states~\citep{li2024error}.

Video diffusion adds temporal conditioning~\citep{ho2022videodiffusion}; large latent or transformer models scale it to high-resolution generation~\citep{blattmann2023svd,yang2025cogvideox,wan2025wan}. Camera-controlled systems inject pose or 3D guidance for viewpoint control and consistency~\citep{he2024cameractrl,ren2025gen3c}. Diffusion world models further condition on camera trajectories, geometry or depth, and history to generate videos that blend seamlessly with the real world~\citep{peebles2023dit,zhu2025aether}.

Broader world-model research learns compact predictive states for simulation and control~\citep{ha2018worldmodels}, including latent imagination in Dreamer~\citep{hafner2019dreamer} and latent trajectory optimization in TD-MPC2~\citep{hansen2024td}. Although their objectives differ from generative video world models, they motivate compact representations of environment dynamics.

\subsection{Diffusion Feature and Representation Caching}

Diffusion caching exploits redundancy at several granularities. DeepCache and Block Caching reuse intermediate features or blocks~\citep{ma2024deepcache,wimbauer2024blockcaching}; ToCa and DuCa selectively update spatial or token states~\citep{zou2024toca,zou2024duca}; TeaCache and EasyCache reuse denoising outputs or transformation vectors~\citep{liu2025teacache,zhou2025easycache}; and TaylorSeer and HiCache forecast feature trajectories from short histories~\citep{liu2025taylorseer,feng2026hicache}. Despite operating in different representation spaces, these methods generally determine cacheability from local feature, token, output, or trajectory changes. However, they may miss how the local defect propagates through later denoising stages and greatly ruins the final output.

For diffusion world models, WorldCache adapts token predictors to local dynamics and resumes exact computation when accumulated token drift grows~\citep{feng2026worldcache}. HERO instead exploits layer-dependent temporal redundancy by selectively refreshing temporally variable patches and extrapolating more stable deep features~\citep{song2025hero}. In contrast, WorldDynCache directly approximates the composed diffusion latent transition and controls its use according to the predicted and accumulated transition risk. Its lifted representation is loosely inspired by the observable-based view of Koopman modeling, from the original linear-operator formulation for nonlinear dynamics to finite-dimensional invariant-subspace and learned-observable approximations~\citep{koopman1931hamiltonian,brunton2016koopman,williams2015koopman,lusch2018deepkoopman}. WorldDynCache, however, uses only a fixed non-learned map and does not identify or assume a globally valid Koopman operator.

\begin{figure}[t!]
\centering
\includegraphics[width=\columnwidth]{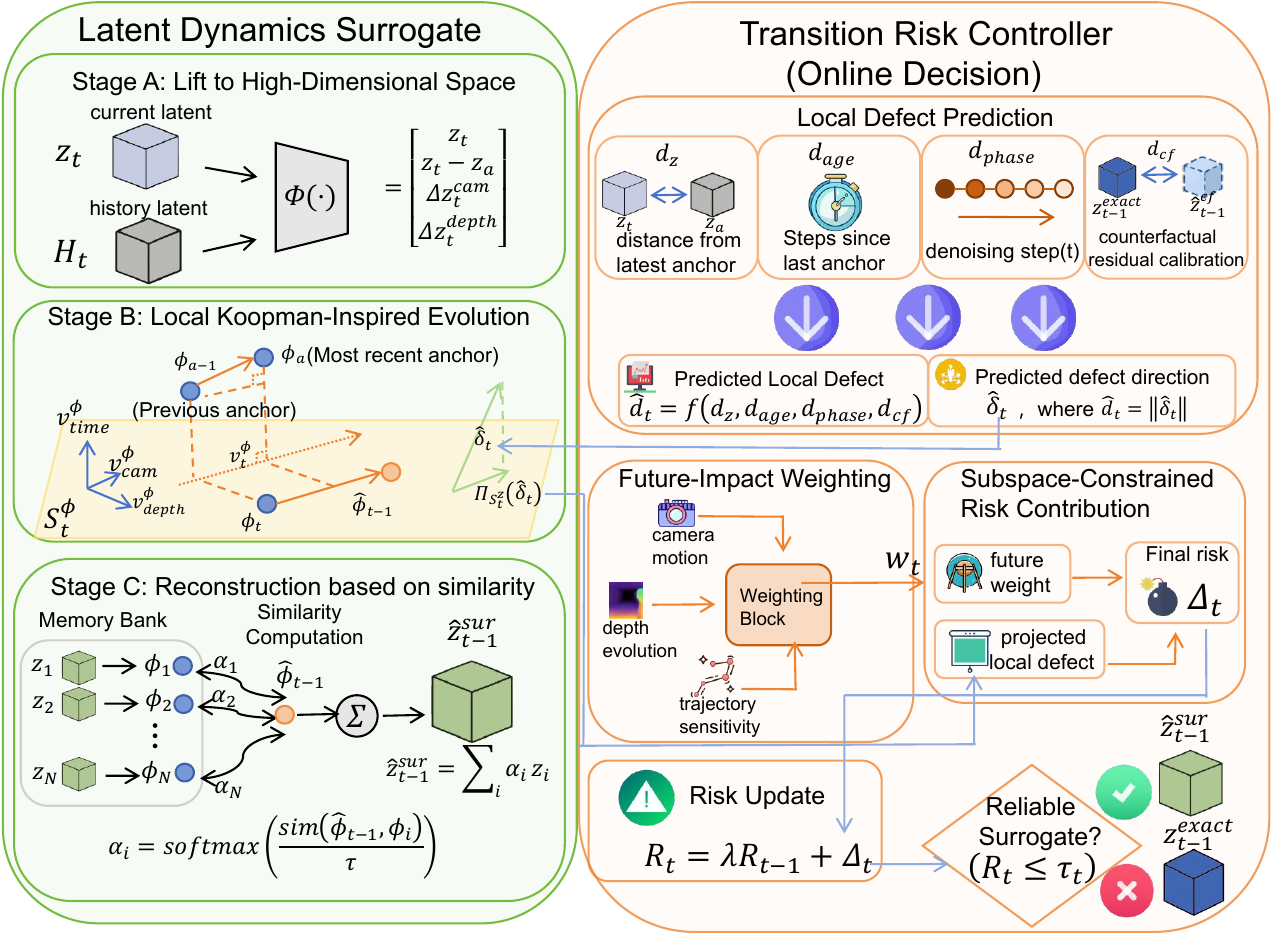}
\caption{Overview of WorldDynCache. The latent dynamics surrogate (green) lifts the current state, applies trajectory-local Koopman-inspired evolution, and reconstructs a candidate next latent from exact-anchor memory. The risk controller (orange) predicts and accumulates its future-weighted transition risk. A low-risk candidate is committed; otherwise, WorldDynCache executes an exact transition, which refreshes memory and calibrates subsequent risk estimates.}
\label{fig:worlddyncache-overview}
\end{figure}

\section{Motivation and Observations}

\subsection{Observation 1: Instantaneous transition defects do not
fully reflect their downstream consequences.}
Our first insight is that the reliability of a surrogate transition
cannot be determined solely by its current defect, because the
resulting perturbation is subsequently evolved by the remaining
denoising transitions. To test this insight, we inject a single
surrogate transition from an exact latent state and then resume exact
inference, measuring both its local defect $d_t$ and downstream
rollout impact $D_t^{\mathrm{down}}$.
Figure~\ref{fig:transition-safety-gap}(a) reveals false-safe cases
whose local defects remain below $\tau_d$ while their downstream
impacts exceed $\tau_D$. Panel~(b) shows that this problem becomes
more pronounced under stronger camera--depth dynamics: the instantaneous-defect criterion produces
false-safe rates of 5.2\%, 14.6\%, and 20.9\% in the Low, Medium,
and High regimes, respectively. These results motivate our
Transition Risk Controller, which weights the predicted defect by
phase and condition sensitivity and accumulates risk across
consecutive surrogate steps. The resulting future-weighted criterion
reduces the corresponding false-safe rates to 3.8\%, 5.4\%, and
9.5\%, demonstrating more reliable decisions particularly under
strong dynamics.

\subsection{Observation 2: Short native-space histories do not reliably
preserve latent transition directions.}
Our second insight is that a displacement estimated from recent
native-space states may cease to predict the next transition direction
as denoising phase and world conditions evolve, with the mismatch
compounding over consecutive surrogate steps. To test this insight,
we measure the angular disagreement between approximate and exact
latent increments from the same current state and compare different
approximation strategies over increasing cache-streak lengths.
Figure~\ref{fig:transition-safety-gap}(c) shows that native-space
reuse and extrapolation accumulate increasingly large direction
errors as the streak grows. A lifted representation reduces this
misalignment, and incorporating phase and condition information
provides a further improvement. These findings motivate our
condition- and phase-aware Latent Dynamics Surrogate, which retrieves
compatible exact transition increments in the lifted space.
WorldDynCache consequently maintains the lowest direction error
across the evaluated cache streaks.

\section{Method}

\subsection{Overview}

WorldDynCache approximates a complete latent transition rather than an intermediate denoising output. Let $H_t$ denote the recent exact-anchor history available at denoising step $t$. The exact and surrogate transitions are
\begin{equation}
\begin{aligned}
z_{t-1}^{\mathrm{exact}}
&=\Psi_t\!\left(z_t,F_\theta(z_t,t,c)\right),\\
\hat z_{t-1}
&=G_t(z_t,H_t).
\end{aligned}
\label{eq:exact-surrogate-transition}
\end{equation}
Here $F_\theta$ is the expensive transformer denoising dynamics, $c$ is the model condition, and $\Psi_t$ is the scheduler transition. Thus, WorldDynCache approximates the composed world-state transition $z_t\rightarrow z_{t-1}$; it does not directly approximate the denoising output $y_t=F_\theta(z_t,t,c)$.

Figure~\ref{fig:worlddyncache-overview} summarizes the inference pipeline. The left green block constructs a candidate $\hat z_{t-1}$ with the \emph{Latent Dynamics Surrogate}; the right orange \emph{Transition Risk Controller} then decides whether to commit it. An accepted candidate becomes the next sampler state; a rejected candidate triggers the original transformer and scheduler. Exact transitions also expose counterfactual surrogate defects that calibrate later decisions. In summary, the surrogate answers how to approximate the next latent, and the controller answers when that approximation should be trusted.

\subsection{Latent Dynamics Surrogate}

Motivated by Observation~2, the surrogate predicts the transition direction in a representation that exposes condition- and phase-dependent world dynamics, rather than extrapolating native-space denoising outputs. The left green block of Figure~\ref{fig:worlddyncache-overview} therefore lifts the current state, evolves it with a trajectory-local approximation, and maps it back to latent space through a nonlinear memory operator.

\begin{figure}[t!]
\centering
\includegraphics[width=\columnwidth]{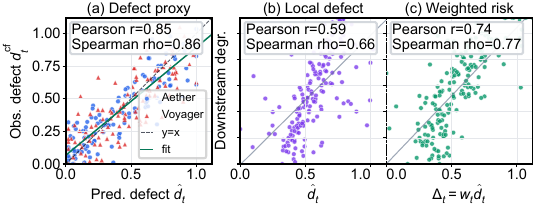}
\caption{Latent-transition risk diagnostics. Panel (a) compares $\hat d_t$ with $d_t^{\mathrm{cf}}$, with Pearson/Spearman correlations of 0.85/0.86. Panels (b,c) compare downstream degradation with the local defect $\hat d_t$ and future-weighted risk $\Delta_t$.}
\label{fig:latent-risk-diagnostic}
\end{figure}

\subsubsection{Condition-Aware Lifted State}

Koopman-inspired modeling suggests that nonlinear state evolution may admit a simpler local approximation when expressed through suitable observables~\citep{williams2015koopman,lusch2018deepkoopman}. Let $z_a$ be the latest exact anchor and let $H_t$ contain recent exact-transition tuples. WorldDynCache constructs
\begin{equation}
\begin{aligned}
\phi_t=\Phi(z_t,H_t)
={}&
\big[
\nu(z_t),\;
\nu(z_t-z_a),\;
\Delta z_{\mathrm{cam}},\\
&\phantom{\big[}
\Delta z_{\mathrm{depth}},\;
q_t,\;
h(H_t)
\big].
\end{aligned}
\label{eq:lifted-state}
\end{equation}
The blocks are the RMS-normalized current world latent and anchor-relative displacement, normalized camera/motion and depth/geometry descriptors, normalized scheduler phase $q_t$, and a summary $h(H_t)$ of recent exact transitions. Unavailable condition descriptors are set to zero, and normalized timestep replaces unavailable scheduler phase. The map $\Phi$ is fixed and non-learned; normalization details and history summaries are specified in the supplementary material.

\subsubsection{Local Lifted Evolution}

Alongside $H_t$, we maintain a unified joint exact-transition bank $\mathcal B_t$. Each record stores the lifted input and output, their increment, the exact output latent, and the corresponding scheduler phases. The bank is updated only after an exact transition under FIFO replacement; surrogate transitions are never inserted. Lifted-transition retrieval uses the input/lifted-increment view of $\mathcal B_t$, while kernelized observation uses its exact-output view.

At the current step, WorldDynCache retrieves only records whose lifted inputs are locally compatible with $\phi_t$, forming $\mathcal N_t\subseteq\mathcal B_t$. The transition-retrieval score and normalized weight are
\begin{equation}
\begin{aligned}
\ell_i
&=
\operatorname{sim}\!\big(s(\phi_t),s(\phi_i^{\mathrm{in}})\big)
-\gamma_q\big\|q_t-q_i^{\mathrm{in}}\big\|_2^2,\\
\pi_i
&=
\frac{\exp(\ell_i/\tau_{\mathrm{in}})}
{\sum_{j\in\mathcal N_t}\exp(\ell_j/\tau_{\mathrm{in}})}.
\end{aligned}
\label{eq:transition-retrieval}
\end{equation}
Here $\pi_i$ is the normalized transition-retrieval weight: higher state/condition similarity increases $\pi_i$, while larger scheduler-phase mismatch decreases it. The scheduler scaling and lifted-state evolution are
\begin{equation}
\begin{aligned}
\alpha_t
&=
\operatorname{clip}\!
\left(
\frac{|q_{t-1}-q_t|}
{|q_{a-1}-q_a|+\epsilon},
\alpha_{\min},
\alpha_{\max}
\right),\\
\hat\phi_{t-1}
&=
\phi_t
+
\alpha_t
\sum_{i\in\mathcal N_t}
\pi_i\Delta\phi_i.
\end{aligned}
\label{eq:lifted-evolution}
\end{equation}
Thus $\pi_i$ selects compatible historical transition directions, and $\alpha_t$ rescales their magnitude to the current scheduler interval; $a\rightarrow a-1$ denotes the latest exact transition interval. Memory capacity, compatibility thresholds, temperatures, clip bounds, and fallback details are provided in the supplementary material. The update is used only over a recent-anchor-supported short horizon and is a trajectory-conditioned, Koopman-inspired local approximation rather than an identified or globally learned Koopman operator.

\subsubsection{Kernelized Nonlinear Observation}

WorldDynCache maps the predicted lifted state back to the next world latent using the exact-output view of the same bank $\mathcal B_t$. For notation, let $\mathcal M_t$ denote the corresponding set of exact-output pairs $(\phi_i^{\mathrm{out}},z_i^{\mathrm{out}})$; $\mathcal M_t$ is not a separately maintained memory. A fixed 256-dimensional summary $s(\cdot)$ supports lightweight retrieval with temperature $\tau$:
\begin{equation}
\begin{aligned}
\bar\kappa_i
&=
\frac{
\exp\!\big(
\operatorname{sim}(s(\hat\phi_{t-1}),s(\phi_i^{\mathrm{out}}))/\tau
\big)
}{
\sum_j
\exp\!\big(
\operatorname{sim}(s(\hat\phi_{t-1}),s(\phi_j^{\mathrm{out}}))/\tau
\big)
},\\
\hat z_{t-1}
&=
\sum_i
\bar\kappa_i z_i^{\mathrm{out}}.
\end{aligned}
\label{eq:kernel-observation}
\end{equation}
Because $\bar\kappa_i$ depends on $s(\hat\phi_{t-1})$, the observation is query-dependent and nonlinear rather than a fixed linear projection. It uses no learned decoder and evaluates no transformer. Equations~\eqref{eq:lifted-state}--\eqref{eq:kernel-observation} define $G_t$ in Eq.~\eqref{eq:exact-surrogate-transition}. A fixed linear readout is evaluated as an ablation in the supplementary material.

\begin{figure*}[t!]
\centering
\includegraphics[width=\textwidth]{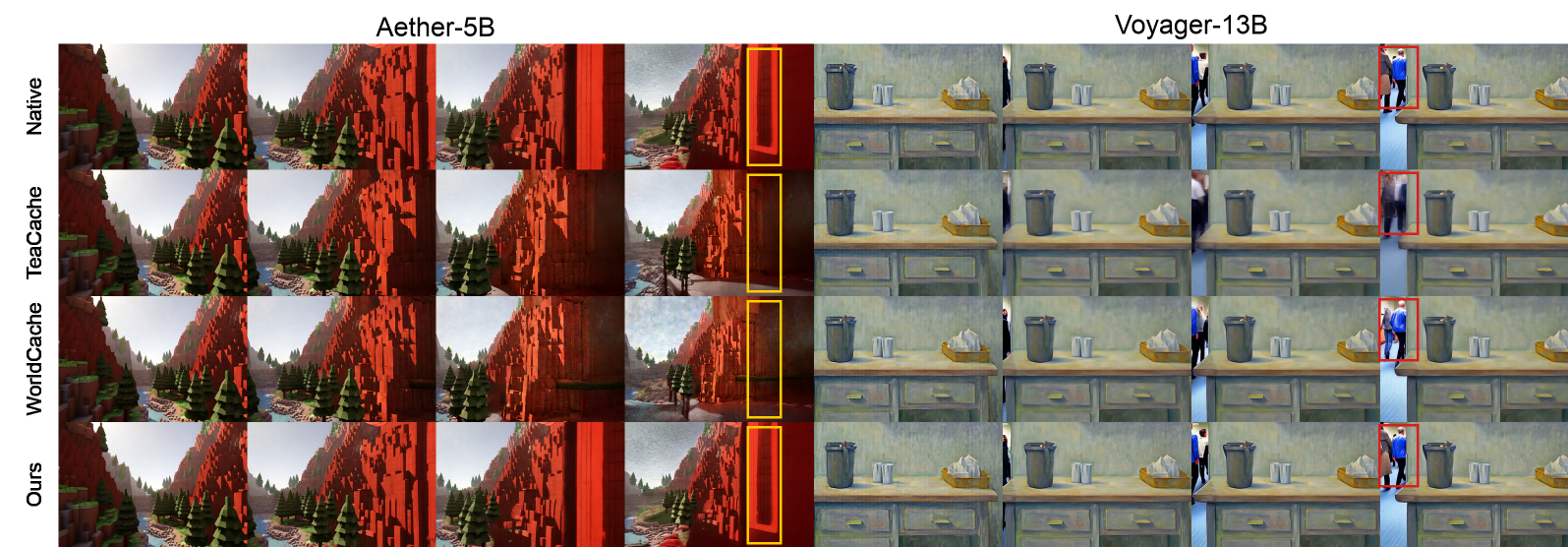}
\caption{Qualitative world-generation comparisons. Left: Aether-5B verdant-landscape sample 091. Right: Voyager-13B stylized indoor-workspace sample. Rows show Native, TeaCache, WorldCache, and WorldDynCache; each row contains four frames from the same input scene. The boxes highlight representative regions where WorldDynCache better preserves the native output.}
\label{fig:qualitative-comparisons}
\end{figure*}

\subsection{Transition Risk Controller}

Motivated by Observation~1, a decision should not depend on the instantaneous local defect alone: similar defects can have very different remaining-rollout consequences under different phases and camera/geometry changes, and their effects can persist across cache streaks. The controller therefore predicts the local defect of the candidate, weights it by future sensitivity, and accumulates the resulting risk. The right orange block of Figure~\ref{fig:worlddyncache-overview} summarizes this procedure.

\paragraph{Calibrated Local Defect Prediction.}
At a candidate surrogate step the exact next latent is unavailable, so WorldDynCache predicts the normalized local defect from inexpensive pre-decision signals:
\begin{equation}
\begin{aligned}
\hat d_t
&=
\eta_d p_t,\\
p_t
&=
\omega_z d_z(t)
+\omega_{\mathrm{age}}d_{\mathrm{age}}(t)
+\omega_{\mathrm{phase}}d_{\mathrm{phase}}(t)\\
&\quad
+\omega_{\mathrm{cam}}d_{\mathrm{cam}}(t)
+\omega_{\mathrm{depth}}d_{\mathrm{depth}}(t).
\end{aligned}
\label{eq:local-defect-proxy}
\end{equation}
Here $d_z$ is the distance from the latest exact anchor, $d_{\mathrm{age}}$ counts consecutive surrogate steps, $d_{\mathrm{phase}}$ measures scheduler-phase displacement, and $d_{\mathrm{cam}}$/$d_{\mathrm{depth}}$ summarize camera/motion and depth/geometry change. Missing condition signals contribute zero; fixed nonnegative weights and normalizations are given in the supplementary material.

Exact transitions calibrate this proxy. At an exact step, the pre-insertion surrogate produces the counterfactual candidate $z_{t-1}^{\mathrm{cf}}=G_t(z_t,H_t)$. The observed counterfactual defect is
\begin{equation}
d_t^{\mathrm{cf}}
=
\frac{
\operatorname{mean}
\big|
z_{t-1}^{\mathrm{exact}}
-
z_{t-1}^{\mathrm{cf}}
\big|
}{
\operatorname{mean}
\big|
z_{t-1}^{\mathrm{exact}}
\big|
+
\epsilon
}.
\label{eq:counterfactual-defect}
\end{equation}
The online calibration update is
\begin{equation}
\eta_d
\leftarrow
\operatorname{clamp}
\!\left(
\beta_d\eta_d
+
(1-\beta_d)
\frac{d_t^{\mathrm{cf}}}{p_t+\epsilon},
\eta_{\min},
\eta_{\max}
\right).
\label{eq:online-calibration}
\end{equation}
The counterfactual query uses pre-insertion history after the exact result is available, adding no transformer evaluation and preventing self-retrieval. Figure~\ref{fig:latent-risk-diagnostic}(a) validates $\hat d_t$ against $d_t^{\mathrm{cf}}$ at exact anchors. On pooled Aether and Voyager transitions, Pearson/Spearman correlations of 0.85/0.86 show strong linear and rank agreement, while concentration near the identity line suggests approximate magnitude calibration. This supports online candidate screening, but by Observation~1, instantaneous defect alone cannot characterize downstream consequences; Figures~\ref{fig:latent-risk-diagnostic}(b,c) test this limitation.

\begin{table*}[t!]
\caption{World generation on HunyuanVoyager-13B (left) and Aether-5B (right). WS-S/WS-D denote WorldScore-Static/Dynamic; Mem is peak memory in GB. The best result among accelerated methods is bold.}
\label{tab:world-generation}
\centering
\begin{minipage}[t]{0.5\textwidth}
\centering
\scriptsize
\textbf{HunyuanVoyager-13B} ($512\times768$, 49 frames)\\[2pt]
\setlength{\tabcolsep}{1.4pt}
\tiny
\resizebox{\linewidth}{!}{%
\begin{tabular}{lccccccc}
\toprule
Method & WS-S$\uparrow$ & WS-D$\uparrow$ & PSNR$\uparrow$ & SSIM$\uparrow$ & LPIPS$\downarrow$ & Speed$\uparrow$ & Mem$\downarrow$ \\
\midrule
Voyager & 66.28 & 46.40 & $\infty$ & 1.000 & 0.000 & 1.00$\times$ & 50.44 \\
DuCa & 53.87 & 37.71 & 16.66 & 0.508 & 0.486 & 1.30$\times$ & 109.70 \\
ToCa & 47.49 & 33.24 & 15.51 & 0.409 & 0.558 & 1.01$\times$ & 107.35 \\
TaylorSeer & 62.46 & 43.72 & 18.32 & 0.615 & 0.293 & 0.88$\times$ & 163.79 \\
HiCache & 63.80 & 44.66 & 18.56 & 0.623 & 0.281 & 0.96$\times$ & 163.79 \\
TeaCache & 60.88 & 42.61 & 16.25 & 0.565 & 0.372 & 3.38$\times$ & 56.52 \\
EasyCache & 64.16 & 44.91 & 21.76 & 0.737 & 0.208 & 3.58$\times$ & 50.98 \\
HERO & 62.37 & 43.67 & 17.71 & 0.601 & 0.315 & 0.96$\times$ & 173.42 \\
WorldCache & 64.89 & 45.43 & 23.49 & 0.770 & 0.176 & 3.65$\times$ & 50.58 \\
\textbf{WorldDynCache} & \textbf{65.23} & \textbf{45.78} & \textbf{25.61} & \textbf{0.812} & \textbf{0.158} & \textbf{4.92$\times$} & 50.48 \\
\bottomrule
\end{tabular}}
\end{minipage}%
\begin{minipage}[t]{0.5\textwidth}
\centering
\scriptsize
\textbf{Aether-5B} ($480\times720$, 41 frames)\\[2pt]
\setlength{\tabcolsep}{1.4pt}
\tiny
\resizebox{\linewidth}{!}{%
\begin{tabular}{lccccccc}
\toprule
Method & WS-S$\uparrow$ & WS-D$\uparrow$ & PSNR$\uparrow$ & SSIM$\uparrow$ & LPIPS$\downarrow$ & Speed$\uparrow$ & Mem$\downarrow$ \\
\midrule
Aether & 64.60 & 45.22 & $\infty$ & 1.000 & 0.000 & 1.00$\times$ & 46.58 \\
DuCa & 60.17 & 42.12 & 26.68 & 0.838 & 0.151 & 1.63$\times$ & 61.44 \\
ToCa & 60.15 & 42.11 & 26.68 & 0.839 & 0.151 & 1.62$\times$ & 61.78 \\
TaylorSeer & 57.11 & 39.97 & 22.92 & 0.713 & 0.324 & 1.66$\times$ & 77.32 \\
HiCache & 58.96 & 41.27 & 24.93 & 0.784 & 0.226 & 1.65$\times$ & 77.32 \\
TeaCache & 60.95 & 42.67 & 26.60 & 0.843 & 0.138 & 1.57$\times$ & 46.78 \\
EasyCache & 62.89 & 44.02 & 22.84 & 0.720 & 0.186 & 1.49$\times$ & 46.59 \\
HERO & 58.62 & 41.04 & 23.56 & 0.741 & 0.259 & 1.36$\times$ & 75.08 \\
WorldCache & 63.68 & 44.72 & 31.87 & 0.924 & 0.066 & 1.68$\times$ & 46.59 \\
\textbf{WorldDynCache} & \textbf{64.32} & \textbf{45.07} & \textbf{33.21} & \textbf{0.955} & \textbf{0.061} & \textbf{2.15$\times$} & 46.59 \\
\bottomrule
\end{tabular}}
\end{minipage}
\end{table*}

\paragraph{Future-Impact-Aware Risk Accumulation.}
The same local defect can be more consequential at a sensitive denoising phase or under substantial camera/geometry change. So WorldDynCache forms the future-sensitivity weight, current risk contribution, and accumulated candidate risk jointly:
\begin{equation}
\begin{aligned}
w_t
&=
\min\!\Big\{
2,\,
1
+\alpha_{\mathrm{cam}}u_{\mathrm{cam}}(t)
+\alpha_{\mathrm{depth}}u_{\mathrm{depth}}(t)\\
&\qquad
+\alpha_{\mathrm{phase}}u_{\mathrm{phase}}(t)
\Big\},\\
\Delta_t
&=
w_t\hat d_t,
\qquad
R_t^{\mathrm{cand}}
=
\lambda R+\Delta_t.
\end{aligned}
\label{eq:risk-accumulation}
\end{equation}
Here $w_t$ is an empirical future-sensitivity weight, $\Delta_t$ is the current future-weighted risk contribution, and $R$ carries decayed risk inherited from earlier surrogate steps. The features $u_{\mathrm{cam}}$, $u_{\mathrm{depth}}$, and $u_{\mathrm{phase}}$ summarize camera motion, depth/geometry evolution, and phase sensitivity; missing signals contribute zero, and their precise definitions appear in the supplementary material.

Figures~\ref{fig:latent-risk-diagnostic}(b) and
\ref{fig:latent-risk-diagnostic}(c) compare downstream degradation with the unweighted local defect and with the future-weighted increment. Future weighting raises the Pearson/Spearman correlations from 0.59/0.66 to 0.74/0.77. Eq.~\eqref{eq:risk-accumulation} accumulates over a surrogate streak, several individually moderate defects can jointly trigger exact fallback.

\paragraph{Reliability Decision.}
The Boolean $\mathrm{Available}_t$ indicates whether the surrogate has valid, sufficiently supported exact-anchor memory, a non-expired anchor of pre-decision age at most six, compatible scheduler phase, finite candidate values, and adequate retrieval support, and whether the current step lies outside mandatory-exact initialization and refinement phases. With threshold $\tau_{\mathrm{risk}}$, the controller selects
\begin{equation}
a_t
=
\begin{cases}
\textsc{Surrogate},
&
\mathrm{Available}_t
\ \text{and}\
R_t^{\mathrm{cand}}\le\tau_{\mathrm{risk}},\\
\textsc{Exact},
&
\text{otherwise}.
\end{cases}
\label{eq:reliability-decision}
\end{equation}
After a surrogate decision, the controller sets $R\leftarrow R_t^{\mathrm{cand}}$ and commits $\hat z_{t-1}$. After an exact transition, it decays the stored risk as $R\leftarrow\lambda R$, commits $z_{t-1}^{\mathrm{exact}}$, calibrates $\eta_d$ when a supported pre-insertion counterfactual is available, and refreshes exact-anchor memory. Unsupported candidates always fall back to exact inference. 

\subsection{Inference and Complexity}

At each denoising step, WorldDynCache constructs a lifted candidate, estimates its accumulated transition risk, and either commits the surrogate or falls back to the exact transformer--scheduler transition. The complete inference algorithm is provided in the supplementary material.

WorldDynCache adds only scalar proxy updates, lifted-state construction, small-memory similarity, kernelized latent interpolation, and one cheap counterfactual surrogate on supported exact steps. It requires no extra transformer forward, backward pass, custom kernel, sparse attention, or learned decoder. The surrogate branch bypasses the full transformer as an implementation consequence of directly approximating the composed latent transition; the scientific target remains the world-state evolution $z_t\rightarrow z_{t-1}$.

\section{Experiments}

\subsection{Experimental Settings}

We evaluate world generation on HunyuanVoyager-13B~\citep{huang2025voyager} at $512\times768$ with 49 frames and Aether-5B~\citep{zhu2025aether} at $480\times720$ with 41 frames, and evaluate 3D reconstruction using the Aether protocol. Baselines include DuCa~\citep{zou2024duca}, ToCa~\citep{zou2024toca}, TaylorSeer~\citep{liu2025taylorseer}, HiCache~\citep{feng2026hicache}, TeaCache~\citep{liu2025teacache}, EasyCache~\citep{zhou2025easycache}, HERO~\citep{song2025hero}, WorldCache~\citep{feng2026worldcache}, and the native models~\citep{zhu2025aether}.

Generation metrics are WS-S/WS-D~\citep{duan2025worldscore}, PSNR, SSIM~\citep{wang2004ssim}, and LPIPS~\citep{zhang2018lpips}; following the Aether evaluation protocol~\citep{zhu2025aether}, reconstruction metrics are AbsRel, $\delta_1/\delta_2$, ATE, and RPE-t/RPE-r. We also report speedup and peak memory. 

All the experiments are conducted on eight NVIDIA A6000 48GB GPUs. For world generation,
the target exact-call ratios are 20\% for HunyuanVoyager-13B and 47\% for
Aether-5B. 
Together, these metrics assess perceptual and world-level fidelity for generation, geometric and pose accuracy for reconstruction, and the practical cost of acceleration.

\subsection{World Generation Task}

\begin{figure}[t]
\centering
\includegraphics[width=\columnwidth]{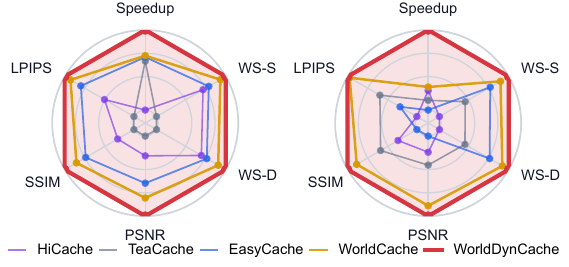}
\caption{Speed--quality tradeoffs for HunyuanVoyager-13B (left) and Aether-5B (right). Each axis is normalized within the corresponding model block among the five plotted methods; larger values are better, with LPIPS inverted.}
\label{fig:generation-speed-quality-radar}
\end{figure}

Table~\ref{tab:world-generation} and Figure~\ref{fig:generation-speed-quality-radar} compare speed and quality on both world models. On HunyuanVoyager-13B, WorldDynCache reaches $4.92\times$ speedup versus WorldCache's $3.65\times$ and improves all reported quality metrics, including WS-S/WS-D from 64.89/45.43 to 65.23/45.78 and LPIPS from 0.176 to 0.158. Its WorldScores remain within 1.05/0.62 points of native Voyager, indicating near-native world-level fidelity.

On Aether-5B, WorldDynCache similarly improves WorldCache's speedup from $1.68\times$ to $2.15\times$, raises WS-S/WS-D from 63.68/44.72 to 64.32/45.07, and reduces LPIPS from 0.066 to 0.061. Its WorldScores are within 0.28/0.15 points of native Aether. Peak memory remains near native inference on both models, indicating that acceleration does not rely on a large auxiliary cache. Relative to the compared caching baselines, these gains yield a more favorable speed--quality tradeoff on both backbones.

Figure~\ref{fig:qualitative-comparisons} shows that WorldDynCache also preserves the main scene layout across the selected rollouts while remaining visually close to native generation.

\subsection{3D Reconstruction Task}

The reconstruction task tests whether acceleration preserves geometry rather than only appearance. Unlike world generation, Aether reconstruction uses a separate task-specific budget. Its realized exact-call ratio is 14\%, corresponding to 7 exact denoiser calls out of 50 denoising steps per rollout. As Table~\ref{tab:reconstruction} shows, WorldDynCache gives the strongest reported depth and pose results while reaching the highest speedup. Compared with WorldCache, AbsRel decreases from 0.341 to 0.337, $\delta_1$ increases from 0.508 to 0.511, and $\delta_2$ increases from 0.741 to 0.768. The trajectory metrics improve concurrently: ATE decreases from 0.184 to 0.174, RPE-t from 0.068 to 0.065, and RPE-r from 0.796 to 0.679. The 14.7\% relative reduction in RPE-r is the largest of these improvements and suggests less accumulated rotational drift.

\begin{table}[t]
\caption{3D reconstruction on Aether-5B. $\delta_1$/$\delta_2$ denote accuracy thresholds $1.25$/$1.25^2$; RPE-t/RPE-r denote translational/rotational RPE; Mem is peak memory in GB.}
\label{tab:reconstruction}
\centering
\resizebox{\columnwidth}{!}{%
\begin{tabular}{lcccccccc}
\toprule
Method & AbsRel$\downarrow$ & $\delta_1\uparrow$ & $\delta_2\uparrow$ & ATE$\downarrow$ & RPE-t$\downarrow$ & RPE-r$\downarrow$ & Spd.$\uparrow$ & Mem$\downarrow$ \\
\midrule
Aether & 0.340 & 0.502 & 0.738 & 0.177 & 0.068 & 0.780 & 1.00$\times$ & 50.19 \\
DuCa & 0.341 & 0.475 & 0.694 & 0.209 & 0.069 & 0.904 & 1.97$\times$ & 52.70 \\
ToCa & 0.341 & 0.476 & 0.694 & 0.209 & 0.069 & 0.904 & 1.98$\times$ & 52.70 \\
TaylorSeer & 0.361 & 0.460 & 0.718 & 0.197 & 0.068 & 1.134 & 2.07$\times$ & 58.57 \\
HiCache & 0.346 & 0.472 & 0.712 & 0.204 & 0.069 & 1.004 & 2.09$\times$ & 58.57 \\
TeaCache & 0.341 & 0.496 & 0.724 & 0.183 & 0.068 & 0.797 & 2.14$\times$ & 50.20 \\
EasyCache & 0.390 & 0.479 & 0.725 & 0.183 & 0.069 & 1.061 & 2.00$\times$ & 50.20 \\
HERO & 0.347 & 0.490 & 0.716 & 0.181 & 0.071 & 0.861 & 1.96$\times$ & 61.56 \\
WorldCache & 0.341 & 0.508 & 0.741 & 0.184 & 0.068 & 0.796 & 2.61$\times$ & 50.20 \\
\textbf{WorldDynCache} & \textbf{0.337} & \textbf{0.511} & \textbf{0.768} & \textbf{0.174} & \textbf{0.065} & \textbf{0.679} & \textbf{3.42$\times$} & 50.20 \\
\bottomrule
\end{tabular}}
\end{table}

Efficiency improves together with geometry: WorldDynCache raises speedup from $2.61\times$ for WorldCache to $3.42\times$, a 31.0\% relative gain, while retaining the same 50.20 GB memory footprint. It also slightly improves every reported reconstruction metric over native Aether. We interpret this result conservatively: the surrogate is not a physical simulator, but risk-controlled latent approximation can avoid the depth and pose degradation seen for more aggressive local reuse methods under this protocol.

\subsection{Ablation Study}

\noindent\textbf{Evaluation protocol.}
We compare all variants under matched surrogate-transition budgets, so they
use approximately the same number of exact transformer calls. We first dissect
the controller and surrogate on Aether-5B, then repeat their factorization on
HunyuanVoyager-13B.

\begin{table}[t]
\caption{Matched-budget component ablation on Aether-5B.}
\label{tab:ablation_aether}
\centering
\scriptsize
\setlength{\tabcolsep}{1.5pt}
\begin{tabular}{@{}p{0.43\columnwidth}cccc@{}}
\toprule
Variant & WS-D$\uparrow$ & LPIPS$\downarrow$ & AbsRel$\downarrow$ & RPE-r$\downarrow$ \\
\midrule
\multicolumn{5}{@{}l}{\textit{Core factorization}} \\
Periodic + native latent secant & 41.23 & 0.236 & 0.365 & 1.029 \\
Risk + native latent secant & 42.45 & 0.124 & 0.348 & 0.836 \\
Periodic + full surrogate & 43.76 & 0.092 & 0.341 & 0.764 \\
\textbf{Full model} & \textbf{45.07} & \textbf{0.061} & \textbf{0.337} & \textbf{0.679} \\
\midrule
\multicolumn{5}{@{}l}{\textit{Surrogate design}} \\
Risk + lifted surrogate, no condition & 43.02 & 0.104 & 0.345 & 0.719 \\
Risk + linear observation & 43.37 & 0.099 & 0.343 & 0.765 \\
\midrule
\multicolumn{5}{@{}l}{\textit{Risk controller}} \\
Instantaneous risk & 44.05 & 0.084 & 0.340 & 0.732 \\
w/o impact weighting & 44.51 & 0.078 & 0.339 & 0.710 \\
w/o CF calibration & 44.24 & 0.089 & 0.337 & 0.691 \\
w/o support constraint & 43.87 & 0.070 & 0.340 & 0.748 \\
\bottomrule
\end{tabular}

\end{table}

\noindent\textbf{Aether component analysis.}
Table~\ref{tab:ablation_aether} yields three findings. (1) In the core
factorization, risk guidance improves the native-sec\-ant baseline from
41.23/0.236 to 42.45/0.124 WS-D/LPIPS, while the lifted surrogate under
periodic placement reaches 43.76/0.092. Combining both performs best at
45.07/0.061 and also gives the lowest AbsRel and RPE-r, confirming that
reliable placement and accurate transition construction are complementary.
(2) Removing condition descriptors or replacing kernelized observation with a
linear readout degrades every metric, validating condition-aware lifting and
nonlinear observation. (3) Using instantaneous risk, removing future-impact
weighting or counterfactual calibration, and removing the support constraint
all reduce WS-D and worsen LPIPS and RPE-r; removing support causes the largest
WS-D drop among controller variants (45.07 to 43.87). AbsRel also worsens
except without counterfactual calibration, where it remains 0.337.

\begin{table}[t]
\caption{Matched-budget controller--surrogate factorization on HunyuanVoyager-13B.}
\label{tab:ablation_voyager}
\centering
\scriptsize
\setlength{\tabcolsep}{1.5pt}
\begin{tabular}{@{}llcccc@{}}
\toprule
Controller & Surrogate & WS-S$\uparrow$ & WS-D$\uparrow$ & PSNR$\uparrow$ & LPIPS$\downarrow$ \\
\midrule
Periodic & Native latent secant & 60.12 & 41.06 & 16.24 & 0.385 \\
Proposed risk & Native latent secant & 62.36 & 43.22 & 17.63 & 0.262 \\
Periodic & Proposed lifted surrogate & 63.09 & 44.17 & 22.46 & 0.234 \\
\textbf{Proposed risk (Full)} & Proposed lifted surrogate & \textbf{65.23} & \textbf{45.78} & \textbf{25.61} & \textbf{0.158} \\
\bottomrule
\end{tabular}
\end{table}

\noindent\textbf{Cross-model validation.}
Table~\ref{tab:ablation_voyager} confirms the same factorization on
HunyuanVoyager-13B: (1) risk guidance improves the native-secant baseline
from 60.12/41.06 to 62.36/43.22 WS-S/WS-D and lowers LPIPS from 0.385 to
0.262, showing that adaptive placement alone benefits both dynamic quality and
perceptual fidelity; (2) the lifted surrogate with periodic placement reaches
63.09 WS-S, 44.17 WS-D, and 22.46 PSNR, demonstrating that more accurate
transition construction remains valuable without adaptive switching; and
(3) their combination again performs best, with 65.23 WS-S, 45.78 WS-D,
25.61 PSNR, and 0.158 LPIPS. This consistent ordering mirrors the Aether
results and indicates complementary rather than redundant gains. Thus, neither
component is sufficient alone, and their interaction generalizes beyond
Aether's denoising schedule.

\FloatBarrier

\section{Conclusion}

We presented WorldDynCache, a training-free framework that approximates composed latent transitions with a condition- and phase-aware lifted surrogate and controls their use through online-calibrated accumulated risk. WorldDynCache achieves $4.92\times$ generation speedup on HunyuanVoyager and $2.15\times$ on Aether with strong fidelity, together with $3.42\times$ acceleration on Aether reconstruction while preserving depth and pose. Ablations confirm the complementary roles of the surrogate and controller. These results establish risk-controlled latent-transition approximation as an effective route to efficient diffusion world-model inference. 

\bibliography{refs}

\end{document}